%% file: main.tex
\documentclass[sigconf]{acmart}
\AtBeginDocument{%
  }

\copyrightyear{2026}
\acmYear{2026}
\setcopyright{cc}
\setcctype{by-nc-nd}
\acmConference[KDD '26]{Proceedings of the 32nd ACM SIGKDD Conference on Knowledge Discovery and Data Mining V.2}{August 09--13, 2026}{Jeju Island, Republic of Korea}
\acmBooktitle{Proceedings of the 32nd ACM SIGKDD Conference on Knowledge Discovery and Data Mining V.2 (KDD '26), August 09--13, 2026, Jeju Island, Republic of Korea}
\acmDOI{10.1145/3770855.3817550}
\acmISBN{979-8-4007-2259-2/2026/08}

\usepackage{multirow}
\usepackage{subcaption}
\usepackage{booktabs}
\usepackage{siunitx}
\usepackage{balance}
\begin{document}
\title{TransXion: A High-Fidelity Graph Benchmark for Realistic Anti-Money Laundering}

\author{Keyang Chen}
\authornote{Both authors contributed equally to this research.}
\orcid{0009-0006-3318-0214}
\affiliation{%
  \institution{Fudan University}
  \city{Shanghai}
  \country{China}
}
\email{chenky25@m.fudan.edu.cn}

\author{Mingxuan Jiang}
\authornotemark[1]
\orcid{0009-0009-0893-7512}
\affiliation{%
  \institution{Fudan University}
  \city{Shanghai}
  \country{China}
}
\email{mxjiang24@m.fudan.edu.cn}

\author{Yongsheng Zhao}
\orcid{0009-0007-4150-6533}
\affiliation{%
  \institution{Fudan University}
  \city{Shanghai}
  \country{China}
}
\email{zhaoys25@m.fudan.edu.cn}

\author{Zeping Li}
\orcid{0009-0007-0099-5879}
\affiliation{%
  \institution{Fudan University}
  \city{Shanghai}
  \country{China}
}
\email{zepingli23@m.fudan.edu.cn}

\author{Zaiyuan Chen}
\orcid{0009-0008-5162-2009}
\affiliation{%
  \institution{Fudan University}
  \city{Shanghai}
  \country{China}
}
\email{zaiyuanchen25@m.fudan.edu.cn}

\author{Weiqi Luo}
\orcid{0009-0005-1090-7903}
\affiliation{%
  \institution{Fudan University}
  \city{Shanghai}
  \country{China}
}
\email{wqluo25@m.fudan.edu.cn}

\author{Zhixin Li}
\orcid{0000-0002-9646-3047}
\affiliation{%
  \institution{Fudan University}
  \city{Shanghai}
  \country{China}
}
\email{lizhixin@fudan.edu.cn}

\author{Sen Liu}
\orcid{0000-0003-2230-7671}
\affiliation{%
  \institution{Fudan University}
  \city{Shanghai}
  \country{China}
}
\email{senliu@fudan.edu.cn}

\author{Yinan Jing}
\orcid{0000-0002-1169-8032}
\affiliation{%
  \institution{Fudan University}
  \city{Shanghai}
  \country{China}
}
\email{jingyn@fudan.edu.cn}

\author{Guangnan Ye}
\authornote{Corresponding authors.}
\orcid{0009-0007-4973-7942}
\affiliation{%
  \institution{Fudan University}
  \city{Shanghai}
  \country{China}
}
\email{yegn@fudan.edu.cn}

\author{Xihong Wu}
\orcid{0009-0004-5236-7469}
\affiliation{%
  \institution{Peking University}
  \city{Beijing}
  \country{China}
}
\email{wxh@cis.pku.edu.cn}

\author{Hongfeng Chai}
\orcid{0000-0002-8577-4771}
\affiliation{%
  \institution{Fudan University}
  \city{Shanghai}
  \country{China}
}
\email{hfchai@fudan.edu.cn}


\renewcommand{\shortauthors}{Keyang Chen et al.}


\begin{abstract}
Money laundering poses severe risks to global financial systems, driving the widespread adoption of machine learning for transaction monitoring. However, progress remains stifled by the lack of realistic benchmarks. Existing transaction-graph datasets suffer from two pervasive limitations: (i) they provide sparse node-level semantics beyond anonymized identifiers, and (ii) they rely on template-driven anomaly injection, which biases benchmarks toward static structural motifs and yields overly optimistic assessments of model robustness.
We propose TransXion, a benchmark ecosystem for Anti-Money Laundering (AML) research that integrates profile-aware simulation of normal activity with stochastic, non-template synthesis of illicit subgraphs.TransXion jointly models persistent entity profiles and conditional transaction behavior, enabling evaluation of "out-of-character" anomalies where observed activity contradicts an entity's socio-economic context. 
The resulting dataset comprises approximately 3 million transactions among 50,000 entities, each endowed with rich demographic and behavioral attributes. Empirical analyses show that TransXion reproduces key structural properties of payment networks, including heavy-tailed activity distributions and localized subgraph structure. Across a diverse array of detection models spanning multiple algorithmic paradigms, TransXion yields substantially lower detection performance than widely used benchmarks, demonstrating increased difficulty and realism. TransXion provides a more faithful testbed for developing context-aware and robust AML detection methods. The dataset and code are publicly available at https://github.com/chaos-max/TransXion.
\end{abstract}




\maketitle

\section{Introduction}
\input{1_Introduction.tex}

\begin{figure*}[t]
  \centering
  \includegraphics[width=\textwidth]{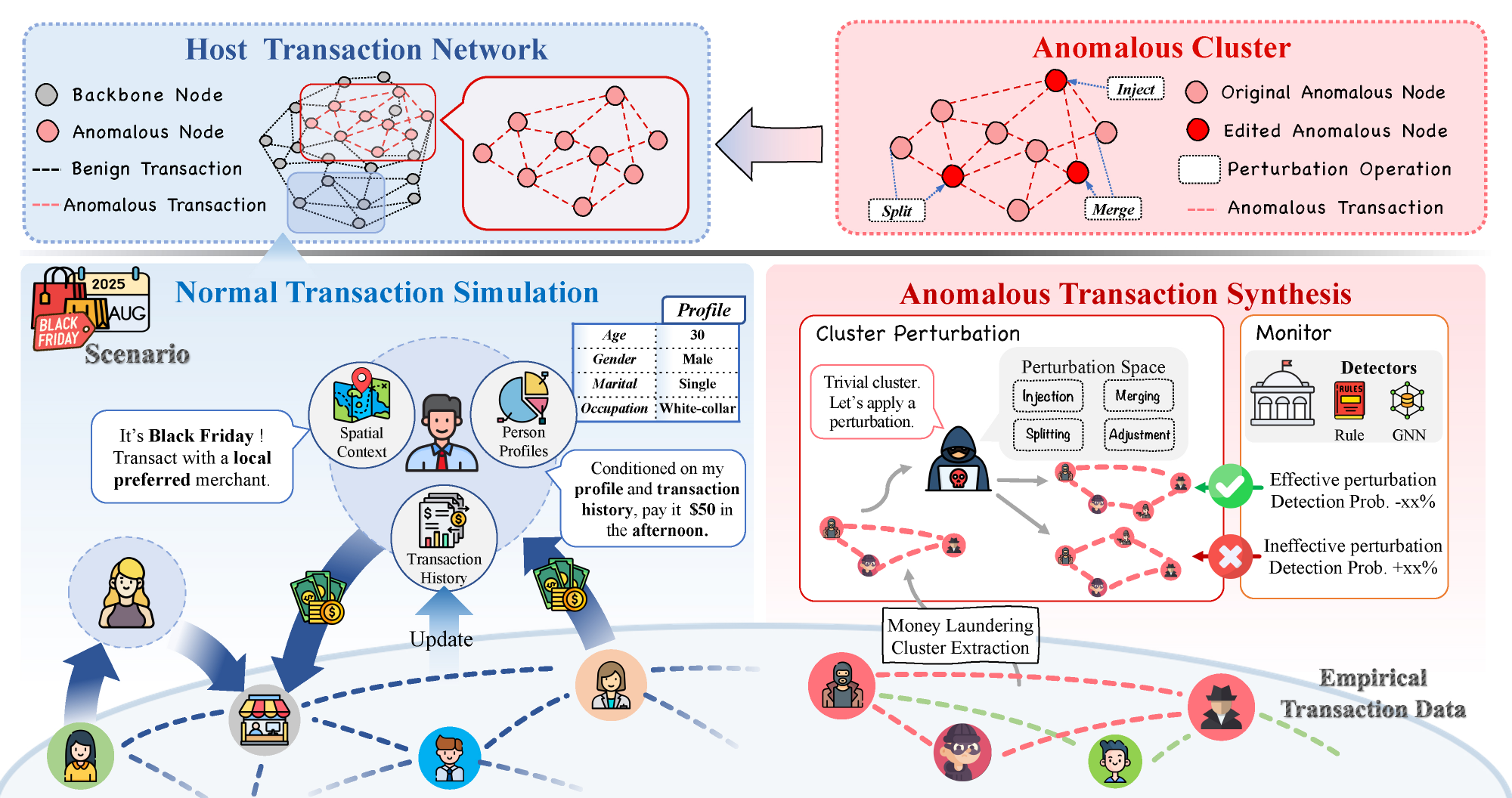}
  \caption{Overview of the TransXion generation framework.
Normal transactions are generated by an agent-based backbone that jointly simulates entity profiles, temporal activity, and network interactions in a closed loop. Realistic illicit behaviors are then synthesized via interpretable perturbations on extracted laundering subgraphs, guided by an adversarial monitor, and embedded back into the normal backbone under structural, profile, and temporal constraints.}
\end{figure*}

\section{Related Work}
\input{2_Related_Work.tex}

\section{Generative Framework}
\input{3_Method.tex}

\section{Proposed Datasets}

\input{4_Datasets.tex}


\begin{table*}[!t]
\centering
\caption{Comparison of graph invariants and tail diagnostics. Invariants are reported as mean $\pm$ std over daily projections. Tail diagnostics use MLE-based fitting with automatically selected $x_{\min}$~\cite{clauset2009power}. \textbf{Strength} and \textbf{Amount} denote node transaction volume and individual transaction values, respectively. Parenthetical values indicate $(x_{\min} \mid n_{\text{tail}})$.}
\label{tab:rq1-fidelity-main}
\fontsize{9pt}{12pt}\selectfont
\begin{tabular*}{\textwidth}{@{\extracolsep{\fill}}lcccc@{}}
\toprule
\textbf{Indicator} & \textbf{Ours} & \textbf{AMLWorld} & \textbf{AMLSim} & \textbf{SAML-D} \\
\midrule
GCC ratio 
& 0.8924 $\pm$ 0.0321
& 0.3824 $\pm$ 0.3373
& 0.8355 $\pm$ 0.0130
& 0.0009 $\pm$ 0.0001 \\
Number of Connected Components
& 308.5 $\pm$ 49.7
& 24{,}102.7 $\pm$ 41{,}639.3
& 389.5 $\pm$ 24.7
& 9{,}624.9 $\pm$ 461.9 \\
Max $k$-core 
& 2.9151 $\pm$ 0.4264
& 1.8333 $\pm$ 0.7638
& 2.93 $\pm$ 0.2551
& 1.0000 $\pm$ 0.0000 \\
Max-core fraction 
& 0.0703 $\pm$ 0.0875
& 0.4581 $\pm$ 0.4456
& 0.0602 $\pm$ 0.0616
& 1.0000 $\pm$ 0.0000 \\
Degree assortativity 
& -0.4943 $\pm$ 0.0279
& -0.2593 $\pm$ 0.2817
& -0.2009 $\pm$ 0.0114
& -0.1188 $\pm$ 0.0048 \\
Global Transitivity
& 0.0021 $\pm$ 0.0005
& 0.0007 $\pm$ 0.0028
& 0.0012 $\pm$ 0.0002
& 0.0000 $\pm$ 0.0000 \\
\midrule
Strength-tail KS $D$ ($x_{\min} \mid n_{\text{tail}}$) 
& 0.0570 \scriptsize{(120.8k $\mid$ 20.0k)} 
& 0.0204 \scriptsize{(531.0M $\mid$ 5.2k)} 
& 0.1046 \scriptsize{(21.5M $\mid$ 4.4k)} 
& 0.0445 \scriptsize{(80.6k $\mid$ 402.1k)} \\
Amount-tail exponent $\alpha$ ($x_{\min} \mid n_{\text{tail}}$) 
& 2.9255 \scriptsize{(40.3k $\mid$ 30.3k)} 
& 1.5423 \scriptsize{(623.8k $\mid$ 253.9k)} 
& 1.3267 \scriptsize{(85.5k $\mid$ 26.5k)} 
& 3.0000$^{\ast}$ \scriptsize{(10.2k $\mid$ 2{,}471.3k)} \\
\bottomrule
\end{tabular*}
\begin{flushleft}
\scriptsize $^{\ast}$ Indicates reaching the boundary of the powerlaw fitting range (typically $\alpha \in [1, 3]$).
\end{flushleft}
\end{table*}

\section{Experiments}
\input{5_Experiments}

\section{Conclusion}

\input{6_Conclusion}

\begin{acks}
This work was supported by the National Key R\&D Program of China (No. 2023YFC3304800) and the National Natural Science Foundation of China (No. 72595845, No. 72595840).
\end{acks}

\clearpage
\bibliographystyle{ACM-Reference-Format}
\balance
\bibliography{ref}

\appendix
\input{7_Appendix}
\clearpage

\end{document}

%% file: 1_Introduction.tex
\begin{figure}[tb]
    \centering
    \includegraphics[width=1.0\linewidth]{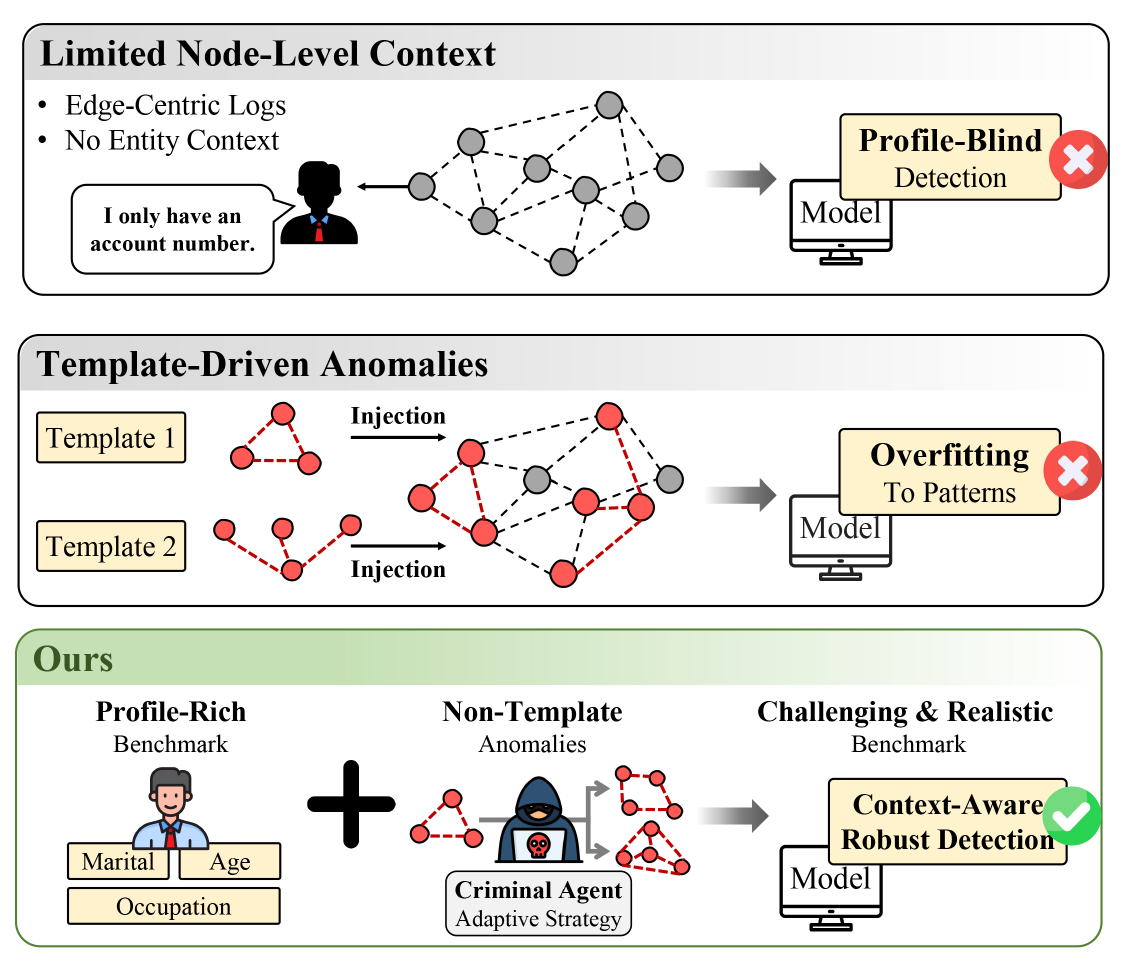}
    \caption{Limitations of prior AML benchmarks versus TransXion. Prior benchmarks use edge-only logs and motif-based injection, whereas \textbf{TransXion} combines profile-rich entities with adaptive anomaly synthesis for contextual detection.}
\end{figure}
Money laundering is the process of disguising the origins of illegally obtained funds to integrate them seamlessly into the global economy\cite{tanzi1996money}. Despite extensive regulatory interventions, the global scale of money laundering remains immense, with conservative estimates situating the volume between 2\% and 5\% of global GDP \cite{reuter2013estimates}. 
This massive scale creates systemic risks beyond individual crimes, which undermines financial integrity, distorts investment decisions, and can destabilize macroeconomic conditions \cite{ferwerda2013effects, walker1999big, unger2006amounts}.

To counter these systemic threats, regulatory bodies and financial institutions are intensifying Anti-Money Laundering (AML) efforts by augmenting traditional rule-based frameworks with machine learning-driven transaction monitoring.
However, real transaction data are rarely shareable at scale due to legal, privacy, and confidentiality constraints, and labels are sparse and incomplete because many laundering activities remain undetected and verification requires expert investigation. As a result, high-fidelity data synthesis has become essential for developing and benchmarking data-driven AML methods \cite{jensen2023synthetic, altman2023realistic}.
In practice, existing benchmarks share a common data organization where transactions are natively recorded as relational event logs. These logs can be cast as a directed multigraph to expose higher-order behavioral patterns exemplified by temporal cascades and localized subgraph motifs\cite{oztas2023enhancing}.
However, existing benchmarks fail to capture the complex integration of semantic context and transactional topology inherent in financial networks, primarily due to two pervasive limitations:

(i) \textbf{Limited Node-Level Context:} 
In operational AML, effective detection depends not only on transactional topology but also on whether behaviors align with customer profiles under Know Your Customer (KYC) standards \cite{Arasa2015DeterminantsOK,fincen2016cddfaq}. 
However, most public benchmarks are constructed as edge-centric event logs, where nodes are either anonymized or limited to categorical identifiers \cite{altman2023realistic}. 
In rare cases where node features exist, they are often embedded representations lacking semantic interpretability, making compliance auditing and model accountability difficult \cite{tang2022rethinking}.
Without structured entity context, models are prone to overlook "out-of-character" behaviors—transactions that deviate from an entity's expected socioeconomic profile but appear plausible in isolation.

(ii) \textbf{Template-Driven Anomaly Injection:} Illicit activity in existing benchmarks is frequently generated by instantiating a restricted set of predefined templates. Under this paradigm, prior works typically synthesize suspicious flows by explicitly manually prescribing sequential laundering stages or by replicating an enumerated set of static graph motifs \cite{altman2023realistic,oztas2023enhancing}. While practical, such reliance on fixed templates restricts benchmarks to predictable patterns. This can overestimate models' robustness: models may perform strongly on these benchmarks yet fail to capture diverse and adaptive laundering strategies encountered in real-world AML.

To address these limitations, we introduce \textbf{TransXion}, a profile-rich transaction-graph benchmark built as an end-to-end synthesis and evaluation ecosystem. TransXion follows a simple three-step pipeline: (i) a normal backbone that jointly generates structured entity profiles and a time-ordered transaction stream; (ii) anomaly synthesis that derives illicit clusters by controlled structural and temporal edits on real illicit subgraphs under budget limits; and (iii) profile-conditioned embedding that inserts each illicit cluster as an intact subgraph by aligning anomalous roles to compatible entities and feasible time windows. The resulting dataset contains roughly \emph{3M transactions} among \emph{50K entities}, with semantically meaningful metadata per entity to support evaluation beyond edge-only logs, including detection of \emph{profile-violating} behaviors. We validate TransXion by matching key stylized properties of payment networks, preserving extreme class imbalance, and demonstrating that template-based baselines can mask performance gaps. Overall, TransXion provides a practical benchmark for developing and stress-testing deployable and auditable AML monitoring under realistic constraints.
Our contributions can be summarized as follows:

\begin{itemize}
    \item \textbf{The First Profile-Rich Transaction Benchmark:} We introduce TransXion, the first large-scale transaction-graph benchmark that integrates approximately 3 million transactions with rich semantically meaningful entity profiles. By providing structured demographics and behavioral metadata, TransXion enables evaluation of AML models conditioned on realistic customer context, filling a critical gap in existing edge-only datasets.
    \item \textbf{Non-Template Anomaly Synthesis:} We propose a novel generation framework that replaces rigid template-driven motifs with perturbation-based synthesis. By injecting diverse anomalies derived from real illicit subgraphs under budget constraints, our framework mitigates benchmark bias toward a set of predefined templates, providing a stronger stress test for detectors.
    \item \textbf{Multi-dimensional Fidelity and Utility Analysis:} We evaluate \textbf{TransXion} along two axes: fidelity, via structural and distributional diagnostics, and utility, via downstream illicit-transaction detection. The results show that \textbf{TransXion} preserves realistic graph signatures while providing a more challenging benchmark than existing alternatives.
\end{itemize}

%% file: 2_Related_Work.tex
\textbf{Public AML datasets and benchmarks.} Real financial transaction data are rarely shareable at scale due to privacy, regulatory, and commercial constraints, and even anonymized records often lack complete and reliable ground truth because many illicit activities remain undiscovered. As noted in prior work, “there are no real public data sets that can be used to investigate and compare anti-money laundering (AML) methods in banks”, and this severe confidentiality barrier limits research on core AML challenges such as extreme class imbalance and model interpretability \cite{jensen2023synthetic}. As a result, only a small number of public benchmarks are available for method development and evaluation.

The Elliptic dataset, a partially labeled Bitcoin transaction graph with around 200K nodes and handcrafted features, has become a widely used benchmark for AML research. However, only a minority of its transactions are annotated as licit or illicit, and the illicit portion is extremely small, which restricts the reliability of supervised evaluation \cite{weber2019anti}. For traditional finance, benchmarks such as T-Finance simulate bank transaction networks with anonymous account nodes described by a limited set of behavioral attributes \cite{tang2022rethinking}. While such datasets enable controlled experimentation, they provide little semantic information about the underlying entities and their socio-economic roles.

More recent public benchmarks have begun to rely on fully synthetic transaction data to increase scale and labeling coverage. For instance, SAML-D provides millions of generated transactions spanning a range of laundering typologies, extending beyond the limited labeled scope of earlier datasets \cite{oztas2023enhancing}. Nevertheless, despite differences in scale and construction, public benchmarks share limitations: they primarily expose transaction-level or topological information, offer sparse or no entity-level semantic context, and abstract away the long-term monitoring setting faced in operational AML systems. Consequently, while these benchmarks are valuable for demonstrating algorithmic feasibility, they remain insufficient to represent the rich entity context and evolving behavioral semantics required for realistic AML risk assessment.

\noindent\textbf{Synthetic data generation for AML.} 
To overcome the limited availability of shareable transaction data and the scarcity of reliable labels, AML research has increasingly relied on simulation and synthesis frameworks to construct controllable, fully annotated transaction networks. Early benchmarks such as PaySim \cite{lopez2016paysim} and AMLSim \cite{weber2018scalable} adopt agent-based simulation: they generate routine customer activities according to predefined behavioral rules and inject canonical laundering motifs (e.g., fan-in/fan-out or smurfing) to create labeled anomalies. This paradigm enables reproducible experimentation under known ground truth and has played an important role in establishing initial baselines.

However, these early generators typically depend on a small set of manually specified templates, restricted transaction types, and simplified institutional assumptions. Such design choices limit the diversity of emergent abnormal patterns that depart from operational banking environments, where laundering strategies are heterogeneous and adaptive rather than fixed. Consequently, performance gains on these benchmarks may primarily reflect a model's ability to recognize injected motifs, rather than its capacity to generalize to unseen or evolving illicit behaviors.

Recent work has sought to improve realism by expanding the synthetic domain or incorporating empirical distributions. For instance, AMLWorld \cite{altman2023realistic} constructs multi-bank, multi-currency virtual environments and simulates a broader range of laundering stages while preserving extreme class imbalance observed in practice. Data-driven approaches such as SynthAML \cite{jensen2023synthetic} fit probabilistic models to real-bank statistics to generate large-scale transaction streams with labeled alerts. These advances substantially enhance fidelity at the level of marginal topology and temporal patterns, and reduce the gap between synthetic and real transaction dynamics.

Despite improved realism, many synthetic benchmarks remain template-centric and evaluate models on predictable injected motifs, which can overstate robustness. In addition, they rarely enforce joint consistency across entity context and temporal behavior, limiting the study of out-of-character laundering. This motivates benchmarks that integrate rich entity profiles with non-template, controllable anomalies and evaluation protocols aligned with operational monitoring.

%% file: 3_Method.tex
\subsection{Normal-Transaction Simulation Backbone}
Most transaction-graph benchmarks release edge-only logs, which removes the entity context needed to evaluate whether activity deviates from an entity's profile. Post hoc attributes cannot recover this coupling because identity and behavior co-evolve through interaction history. We therefore build an agent-based backbone that generates profiles and transactions jointly in a closed loop. Concretely, in each one-hour window we (i) sample initiators, (ii) sample counterparties, (iii) sample event attributes, and (iv) update the interaction state for the next window.

\textbf{Profile-grounded behavior for individual consistency.}
Each entity is initialized with a persistent, semantically meaningful profile that constrains both its activity propensity and attribute ranges. For accounts, profiles include socio-economic attributes and behavioral priors such as daily intensity, typical amount scale, and diurnal rhythms; for merchants, they include business type and operating scale. Operationally, profiles parameterize the per-window activation probability and the attribute samplers, so per-entity behavior stays stable and auditable, while still allowing \emph{out-of-character} deviations to be defined relative to profile expectations.

\textbf{Scenario modulation for macro-level variation.}
To capture low-frequency shifts such as holidays or regional demand changes, we add a lightweight scenario controller that updates at a fixed daily cadence. Each day it outputs a small set of multiplicative modifiers, including a global intensity factor and region-level mixing weights. These modifiers only affect aggregate volume and regional composition; they do not specify individual events or override profile-driven micro behavior. Fixing the update granularity and the affected variables makes the macro layer reproducible and prevents unconstrained tuning.

\textbf{Socially structured sampling for realistic graph topology.}
Within each hourly window, initiators are sampled proportional to profile-derived activity weights to reflect heterogeneous participation. Counterparties are then selected by combining four terms: locality preference, time-decayed interaction memory, merchant attractiveness, and limited exploration for new ties. In minimal form, this is a weighted mixture over candidate counterparties, where the memory term is updated from past interactions and decays with time. This yields heavy-tailed participation and non-trivial clustering patterns, while keeping amounts and other attributes consistent with the initiator profile.

\textbf{Evolving interaction state for temporal stability.}
We maintain an interaction state that summarizes recent history and feeds back into future sampling. The state stores time-decayed partner preferences and region hotspots, and it is updated after each hourly window. This creates path dependence: repeated ties become more likely, but new links still emerge stochastically. As a result, the generated graphs exhibit temporal persistence and stable diurnal structure, alongside heavy-tailed activity patterns observed in empirical financial networks and interbank transaction systems \cite{altman2023realistic}.

\subsection{Anomalous Transaction Synthesis} To generate diverse and adaptive money laundering behaviors, we synthesize anomalies by applying controlled, interpretable edits to realistic illicit transaction subgraphs. Each edit preserves basic transaction sanity, including valid endpoints, non-negative amounts, and a consistent time order. We define a compact action space with four operation families: \textbf{Intermediary Injection} inserts additional intermediate accounts to lengthen transaction chains and obscure direct fund provenance; \textbf{Account Merging} consolidates behaviorally similar accounts to reduce redundant structures and mask repetitive patterns; \textbf{Account Splitting} disperses transaction flows across multiple recipients to dilute concentrated signals that commonly trigger detection heuristics; and \textbf{Transaction Adjustment} perturbs transaction amounts or timestamps within feasible ranges to break exploitable regularities while remaining consistent with surrounding activity. 

While these operations provide the necessary tools, the core challenge lies in selecting the optimal edit sequence to bypass detection. To address this, we frame the synthesis as an adversarial process guided by a pre-trained monitor. The synthesis agent proposes edits for each illicit cluster, while the monitor scores results to encourage modifications that reduce detectability without sacrificing operational plausibility. To bridge the gap between static edits and dynamic strategy, we formulate operation selection as a constrained sequential decision process, refining the agent through a lightweight iterative optimization loop driven by the monitor's feedback. This allows the agent to discover sophisticated, multi-step evasion patterns that are difficult to capture through simple heuristic rules. Further details are in Appendix \ref{Appendix:Anomalous}.

\begin{figure}
    \centering
    \includegraphics[width=1.0\linewidth]{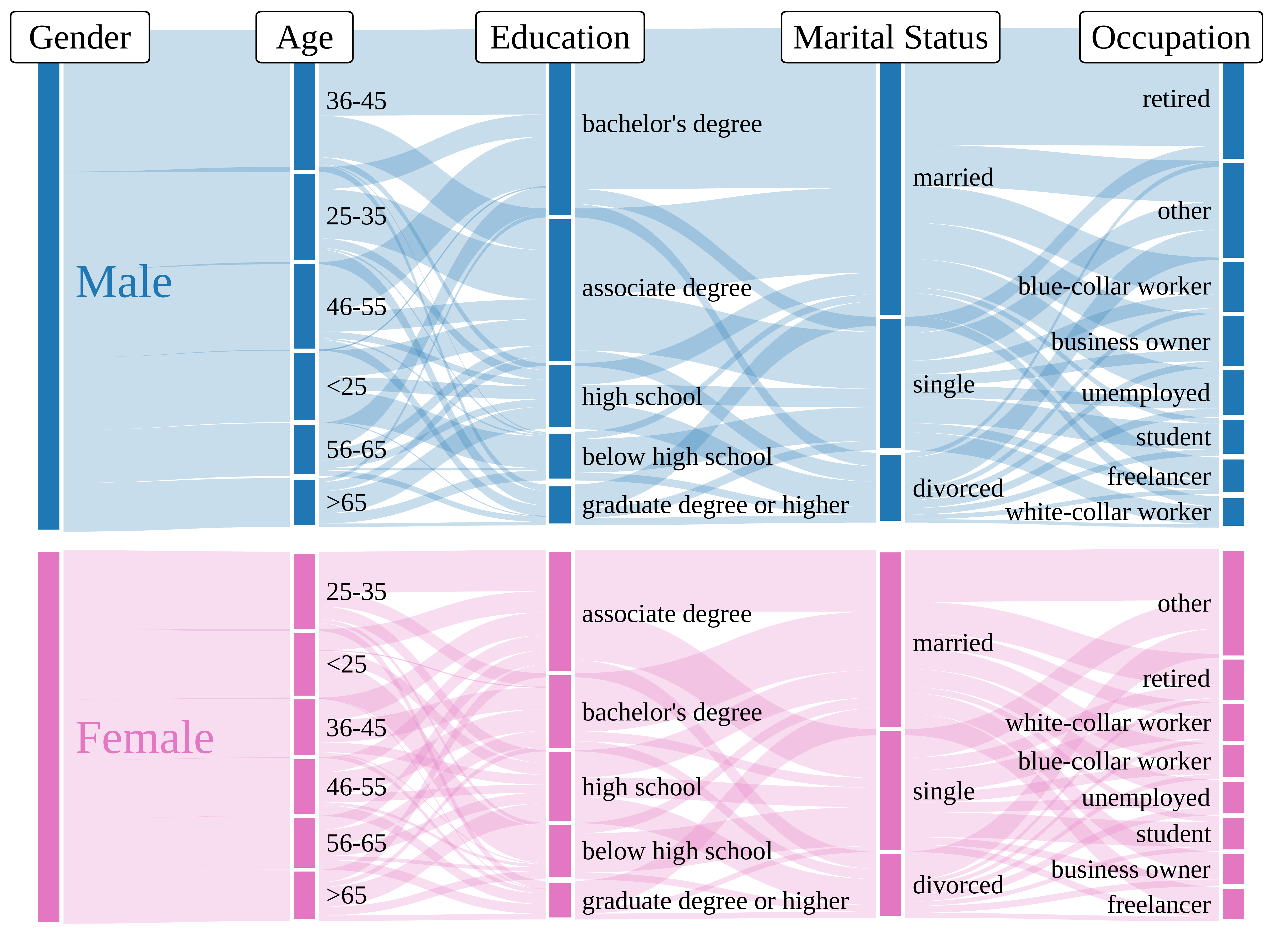}
    \caption{Entity profile co-occurrence in TransXion. Sankey diagram of joint distribution of account-holder attributes.}

    \label{fig:sankey}
\end{figure}

\subsection{Anomaly-to-Normal Embedding}
To prevent injection artifacts, we avoid naive placement and instead embed synthesized clusters as intact subgraphs into the normal backbone. Our objective is to preserve the illicit semantics characterized by internal flow topology and temporal ordering, while ensuring operational feasibility within the normal account space. Crucially, labels are transferred along fund traces during embedding, so that illicitness is preserved by construction.

The embedding is accepted only if it satisfies three constraints: (i) \textbf{Structural Preservation:} preserve all sender--receiver pairs under a one-to-one mapping of cluster participants; (ii) \textbf{Profile Alignment:} match each anomalous role to a normal account whose profile-consistent activity regime falls within the role's required range; (iii) \textbf{Temporal Coherence:} place the cluster into a feasible time window while preserving all intra-cluster sequencing.

We adopt an atomic injection policy, rejecting any cluster that violates these constraints to maintain semantic integrity. At the dataset level, we verify that the augmented backbone preserves key stylized properties, such as heavy-tailed activity statistics. These properties are empirically verified in our experiments. Overall, this embedding balances structural preservation with contextual feasibility, ensuring the benchmark evaluates structure-aware illicit patterns rather than artifacts of injection.

%% file: 4_Datasets.tex
\textbf{The TransXion Dataset}. We introduce \textbf{TransXion}, a relational dataset generated by our framework that contains 3 million transactions among 50{,}000 entities and is organized into \textit{Persons}, \textit{Merchants}, and \textit{Transactions}.TransXion tightly couples transaction-network topology with rich semantic context: the entity tables provide structured profiles linked to unique accounts. Figure~\ref{fig:sankey} visualizes the co-occurrence structure of these profile attributes, illustrating the rich and heterogeneous diversity of account contexts available for learning. Detailed statistical characterization and distribution analyses are provided in Appendix 
\ref{Appednix:detailed dataset}.

%% file: 5_Experiments.tex
This section provides a comprehensive empirical evaluation of \textbf{TransXion} from two complementary perspectives: structural fidelity to stylized facts of transaction networks and utility as a challenging benchmark for illicit-activity detection. Accordingly, we organize our analysis around two research questions.

\noindent\textbf{RQ1 (Fidelity).} To what extent do the statistical properties and graph invariants of \textbf{TransXion} align with patterns documented in real-world payment systems?

\noindent\textbf{RQ2 (Detection Hardness).} To what extent does \textbf{TransXion} increase the difficulty of illicit-transaction detection relative to established benchmark datasets?

\subsection{Experimental Setups}
\label{sec:exp-setup}
\paragraph{Experimental Protocol.}To evaluate both generative fidelity and benchmark utility, we represent the transaction data in two complementary forms. For structural invariants (RQ1), we project the transactions into an undirected simple graph, allowing us to measure stable topological signatures such as connectivity, transitivity, and core-periphery structures. For illicit detection tasks (RQ2), we preserve the full granularity of the data using a directed, weighted temporal multigraph, enabling models to learn from both flow directionality and temporal dynamics.

\textit{Baselines.} We evaluate our benchmark against three established synthetic transaction datasets, each representing a different generation of modeling approaches: 

\noindent\textbf{AMLSim} \cite{weber2018scalable}: A synthetic transaction dataset generated by a multi agent simulation framework. It provides account level transactions together with injected illicit activity patterns and has been used in prior work as a standard synthetic benchmark.

\noindent\textbf{AMLWorld} \cite{altman2023realistic}: An agent-based synthetic financial transaction generator that builds a multi-agent virtual world with both legitimate and criminal agents, producing realistic, standardized AML datasets at large scale and providing complete laundering labels and pattern annotations for benchmarking AML detection models.

\noindent\textbf{SAML-D} \cite{oztas2023enhancing}: A synthetic AML transaction monitoring dataset generated by an enhanced transaction simulation framework. It contains millions of tabular transaction records annotated with 11 normal transaction types and 17 suspicious typologies. 


\subsection{Fidelity of Statistical Properties and Graph Invariants}
Following Soram\"{a}ki et al.~\cite{soramaki2007topology}, we construct daily directed transaction networks and evaluate fidelity by checking whether TransXion reproduces stable regularities of transaction graphs at both the global topology level and the distributional tail level. For comparability across datasets, we compute graph invariants on daily undirected projections and report results as mean $\pm$ std. Furthermore, tail diagnostics are reported on aggregated distributions using a consistent selection rule.

\paragraph{Macro-topology: Backbone Connectivity and Fragmentation.}
Empirical literature show that payment networks are sparse yet dominated by a large connected backbone~\cite{bech2010topology,craig2014interbank}. Motivated by this evidence, we quantify macro-level integration using the giant connected component ratio. As shown in Table~\ref{tab:rq1-fidelity-main}, TransXion preserves an integrated daily topology with a mean GCC ratio of $0.8924 \pm 0.0321$, avoiding the excessive fragmentation observed in some baselines. 
Such structural integrity is important for downstream utility, since it enables large-scale information diffusion and contagion-style propagation over realistic payment backbones~\cite{allen2000financial, acemoglu2015systemic}. 
Accordingly, TransXion provides a more faithful substrate for graph learning, where models operate on globally connected transaction networks rather than fragmented components.

\paragraph{Triadic Closure and Localized Structure}
Real interbank and money-market networks typically exhibit low clustering, yet still show non-trivial local closure attributable to repeated relationships and market segmentation~\cite{boss2004network, bech2010topology}. We quantify this via undirected transitivity, representing the fraction of closed triplets among all connected triplets.
On daily graph projections, TransXion yields a transitivity of $0.0021 \pm 0.0005$, consistent with sparse closure in common benchmarks. This demonstrates that TransXion captures the subtle yet essential signature of localized closure, maintaining the cohesive community structures found in real-world financial interactions. Such structural fidelity yields a meaningful topology in which graph-based models can learn local relational patterns, whereas overly fragmented environments often eliminate these closure signals and render such learning ineffective.

\paragraph{Tiering, Mixing Patterns, and Structural Hierarchy}
A defining feature of financial networks is a tiered architecture, where central institutions intermediate flows for a wide periphery~\cite{anand2018missing}. Such structures are typically characterized by disassortative mixing, where high-degree hubs connect disproportionately to low-degree peripheral nodes. 
Table~\ref{tab:rq1-fidelity-main} shows that TransXion exhibits a pronounced negative assortativity coefficient of $-0.4943$, consistent with hub-and-spoke intermediation~\cite{newman2002assortative}. We further probe hierarchical depth via $k$-core decomposition. \textbf{TransXion} attains a maximum core number of $2.9151$, 
indicating a multi-layered core structure compared to shallower baselines. In contrast to the degenerate 1-core structures found in simpler baselines, \textbf{TransXion} provides the necessary topological depth to evaluate detection models against sophisticated, multi-stage financial interactions.

\paragraph{Heavy Tails and Economic Plausibility of Flows}
Heavy-tailed distributions are ubiquitous in financial systems, particularly for exposure proxies such as node transaction volume and transfer amounts~\cite{gabaix2016power}. We assess tails using MLE-based fitting with an automatically selected lower cutoff $x_{\min}$ following standard practice~\cite{clauset2009power}. 
Under this protocol, TransXion exhibits a low strength-tail KS distance and an amount-tail exponent in a plausible range, indicating that the tail behavior is both heavy-tailed and statistically coherent. Overall, these diagnostics support that TransXion better captures economically grounded flow heterogeneity while avoiding pathological tail behavior.

\paragraph{Summary for RQ1.}
Across macro connectivity, meso-scale tiering, and tail behavior, TransXion exhibits a consistently realistic structural profile that aligns with multiple stylized facts of operational payment networks. Across the compared benchmarks, we observe different emphases and trade-offs in connectivity, core--periphery organization, and tail regimes; in contrast, TransXion maintains a balanced combination of backbone connectivity, hierarchical depth, and economically plausible flow heterogeneity. 
Under this balance, TransXion offers a transaction graph that enables meaningful detection evaluation while retaining the topological richness required for effective graph-based learning.

\begin{table}[!t]
\centering
\caption{Illicit-transaction detection performance using GBT baselines. For each detector, the \textbf{lowest} values are bolded and the \underline{second lowest} values are underlined.}
\label{tab:gbt_results}
\fontsize{9.5pt}{10.5pt}\selectfont
\begin{tabular*}{\columnwidth}{@{\extracolsep{\fill}}ll cc@{}}
\toprule
\multirow{2.5}{*}{\textbf{Dataset}} &
\multirow{2.5}{*}{\textbf{Metric}} &
\multicolumn{2}{c}{\textbf{GBT Detectors}} \\
\cmidrule(lr){3-4}
& & LightGBM & XGBoost \\
\midrule

\multirow{3}{*}{AMLSim}  
  & AUC & \underline{0.4906} & 0.9931 \\
  & AP  & 0.2035 & 0.9206 \\
  & F1  & 0.3598 & 0.9171 \\
\midrule

\multirow{3}{*}{AMLWorld}  
  & AUC & 0.8549 & 0.9631 \\
  & AP  & 0.1369 & 0.3761 \\
  & F1  & \underline{0.2569} & 0.4046 \\
\midrule

\multirow{3}{*}{SAML-D}  
  & AUC & 0.7906 & 0.8521 \\
  & AP  & \underline{0.1361} & \underline{0.2436} \\
  & F1  & 0.2654 & \textbf{0.1872} \\
\midrule

\multirow{3}{*}{\textbf{TransXion}}  
  & AUC & \textbf{0.4128} & \textbf{0.7869} \\
  & AP  & \textbf{0.0313} & \textbf{0.1716} \\
  & F1  & \textbf{0.1383} & \underline{0.2375} \\
\bottomrule
\end{tabular*}
\end{table}

\begin{table*}[!t]
\centering
\caption{Illicit-transaction detection performance comparison. For each detector, the \textbf{lowest} values are bolded and the \underline{second lowest} values are underlined, highlighting the relative difficulty across benchmarks.}
\label{tab:combined_results_gnn}
\fontsize{10pt}{12pt}\selectfont
\begin{tabular*}{\textwidth}{@{\extracolsep{\fill}}ll ccccc@{}}
\toprule
\multirow{2.5}{*}{\textbf{Dataset}} &
\multirow{2.5}{*}{\textbf{Metric}} &
\multicolumn{5}{c}{\textbf{GNN Detectors}} \\
\cmidrule(lr){3-7}
& & Multi-GIN & Multi-GIN+EU & Multi-PNA & Multi-PNA+EU & Multi-GAT \\
\midrule

\multirow{3}{*}{AMLSim}
  & AUC & \textbf{0.929836} & \textbf{0.931760} & \textbf{0.948663} & \textbf{0.940602} & \textbf{0.931399} \\
  & AP  & 0.693337 & 0.728561 & 0.637610 & \underline{0.596686} & 0.458262 \\
  & F1  & 0.775000 & 0.802885 & 0.697337 & \underline{0.700535} & 0.486216 \\
\midrule

\multirow{3}{*}{AMLWorld}
  & AUC & 0.986481 & 0.982223 & 0.982513       & 0.985354 & 0.978961 \\
  & AP  & \underline{0.638040} & \underline{0.504771} & \underline{0.630889}       & 0.672263 & \underline{0.395064} \\
  & F1  & \underline{0.636644} & \underline{0.652040} & \underline{0.627935}       & 0.709499 & \underline{0.451544} \\
\midrule

\multirow{3}{*}{SAML-D}
  & AUC & 0.999975 & 0.999746 & 0.999934 & 0.999984 & 0.999757       \\
  & AP  & 0.991920 & 0.991262 & 0.988842 & 0.994835 & 0.971794       \\
  & F1  & 0.912689 & 0.937154 & 0.968137 & 0.976988 & 0.939498       \\
\midrule

\multirow{3}{*}{\textbf{TransXion}}
  & AUC & \underline{0.968837} & \underline{0.952841} & \underline{0.968291} & \underline{0.968088} & \underline{0.951603} \\
  & AP  & \textbf{0.287842} & \textbf{0.268952} & \textbf{0.350831} & \textbf{0.345992} & \textbf{0.277935} \\
  & F1  & \textbf{0.433980} & \textbf{0.424686} & \textbf{0.486515} & \textbf{0.467446} & \textbf{0.433566} \\
\bottomrule
\end{tabular*}
\end{table*}

\subsection{Detection Hardness and Benchmark Utility}
In this section, we evaluate the downstream utility of \textbf{TransXion} as a rigorous testbed for illicit activity detection. A robust benchmark should present a non-trivial environment where detection models cannot achieve high scores through simple structural cues that do not reflect real-world laundering complexity.

\subsubsection{Hardness Evaluation with Representative Detectors.}
We formulate illicit activity detection as a supervised edge-level classification task. For each dataset, we keep the original event granularity as a directed temporal multigraph, where nodes are accounts and edges are timestamped transactions with associated amounts. We retain parallel edges between the same ordered account pair to capture repeated interactions and temporal dynamics. Since several benchmarks provide no node attributes, we train and evaluate all models using edge features only to ensure fair comparison. Each dataset is chronologically split into train, validation, and test sets with a 6:2:2 ratio by timestamp, ensuring that models are trained on earlier transactions and evaluated on later transactions. We report AUC, AP, and F1 on the held-out test set.

\paragraph{Detection Models.}
To evaluate TransXions hardness, we evaluate various representative detection paradigms, including Gradient Boosted Trees (GBTs) and Graph Neural Networks (GNNs).

\textbf{GBT Detectors.}
We include gradient-boosted tree models as strong tabular baselines, evaluating LightGBM~\cite{ke2017lightgbm} and XGBoost~\cite{chen2016xgboost}. Both are trained as standard binary classifiers directly on per-transaction attributes from the event logs. Since they do not leverage graph structure, they serve as a non-graph reference point for assessing the value of relational and multi-hop signals.

\textbf{GNN Detectors.}
We evaluate graph neural networks for edge-level classification on directed temporal multigraphs. We adopt the directed-multigraph framework of Egressy et al.~\cite{egressy2024provably} and instantiate Multi-GIN~\cite{xu2018powerful} and Multi-PNA~\cite{corso2020principal} backbones, which aggregate incoming and outgoing neighborhoods and support parallel edges. We additionally consider an edge-update variant (+EU)~\cite{battaglia2018relational} and a Multi-GAT~\cite{velivckovic2017graph} baseline for direction-specific attention. For fair comparison, all GNNs are trained using edge-level attributes only.

\paragraph{Main Benchmark Results.}
Tables~\ref{tab:gbt_results} and~\ref{tab:combined_results_gnn} summarize detection performance across tabular and graph-based paradigms on all benchmarks. We highlight two complementary observations: feature-only tabular detectors degrade most on TransXion, and graph-based detectors also show their largest degradation in practical operating performance on TransXion. We analyze these patterns below for GBT and GNN models, respectively.

\textit{GBT: Event-level attributes alone do not support reliable detection under profile-conditioned behavior.}
Table~\ref{tab:gbt_results} shows that gradient-boosted tree baselines trained only on per-transaction attributes degrade most severely on TransXion. In particular, LightGBM and XGBoost attain their weakest performance on TransXion, with AP dropping to 0.0313 and 0.1716, indicating that event-log fields and timestamp-derived cues that work well on prior benchmarks become substantially less discriminative. This drop is unlikely to be explained solely by scale or class imbalance, since feature-only models still perform well across metrics on the other benchmarks.

These observations match our goal of profile-conditioned synthesis: transactions that look similar at the event level can carry different risk depending on the participating accounts. 
Consequently, event-level features alone are less discriminative, and detection benefits from broader contextual signals, as reflected by the gains observed when profile information is introduced in Sec.~\ref{Effect of Node-Level Context}.

\textit{GNN: Hardness concentrates in operating-point performance rather than global ranking.}
Table~\ref{tab:combined_results_gnn} shows a consistent pattern across all GNN detectors: TransXion yields the lowest AP and the lowest F1, while its AUC is the second-lowest among benchmarks. This is expected because AUC evaluates pairwise ranking over all positive--negative pairs and is influenced by separability across the entire score distribution. In contrast, AP and F1 are dominated by the high-score region and by thresholded operating points, which better reflect practical alerting under extreme class imbalance.

This perspective also clarifies why AMLSim attains the lowest AUC while maintaining relatively strong AP and F1. AUC penalizes broad overlap between positive and negative score distributions across the full ranking, leading to frequent mis-orderings among randomly sampled pairs. Meanwhile, AP and F1 can remain favorable when positives are sufficiently concentrated among the highest scores, yielding a high-precision region and a viable operating threshold, even if substantial interleaving persists in the mid- and low-score regions. Therefore, AMLSim may admit reasonable thresholded performance even when the overall score distributions exhibit substantial overlap.

By comparison, TransXion shifts difficulty toward the decision-critical region: illicit edges are synthesized to be behaviorally plausible and context-consistent, which increases ambiguity among high-score candidates and suppresses early precision. Consequently, neighborhood aggregation alone does not recover reliable operating-point performance. Overall, these results indicate that TransXion reduces shortcut learning from template-like motifs and provides a more demanding evaluation setting.

\begin{figure}
    \centering
    \includegraphics[width=1\linewidth]{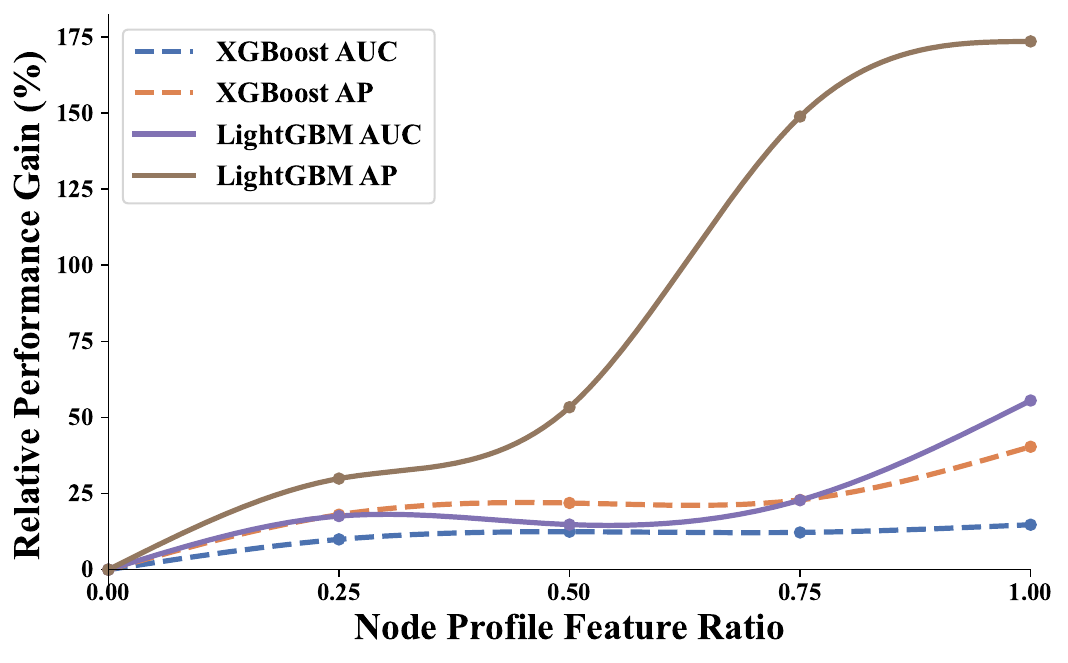}
    \caption{Sensitivity of XGBoost and LightGBM performance to the fraction of node profiles used during training on \textbf{TransXion}, reported in relative AUC and AP change.}
    \label{fig:profile}
\end{figure}
\subsubsection{Effect of Node-Level Context.}
\label{Effect of Node-Level Context}
We investigate whether node-level semantic context provides measurable utility for illicit transaction detection by comparing profile-blind and profile-aware variants of tabular detectors on both TransXion and AMLWorld. Profile-aware models augment transaction features with entity profile attributes, whereas profile-blind models rely solely on edge-level statistics. To assess whether profiles improve a detector's ranking and retrieval of illicit transactions under varying decision thresholds, we focus on AUC and AP as the primary criteria throughout this section.

\textit{Progressive node-context ablation.}
We first perform an ablation on TransXion by gradually increasing the number of available profile attributes used during training. As shown in Fig \ref{fig:profile}, both XGBoost and LightGBM exhibit consistent performance gains as more node attributes are retained, which indicates that node context provides complementary signal beyond transaction-level features.

\begin{figure}
    \centering
    \includegraphics[width=1\linewidth]{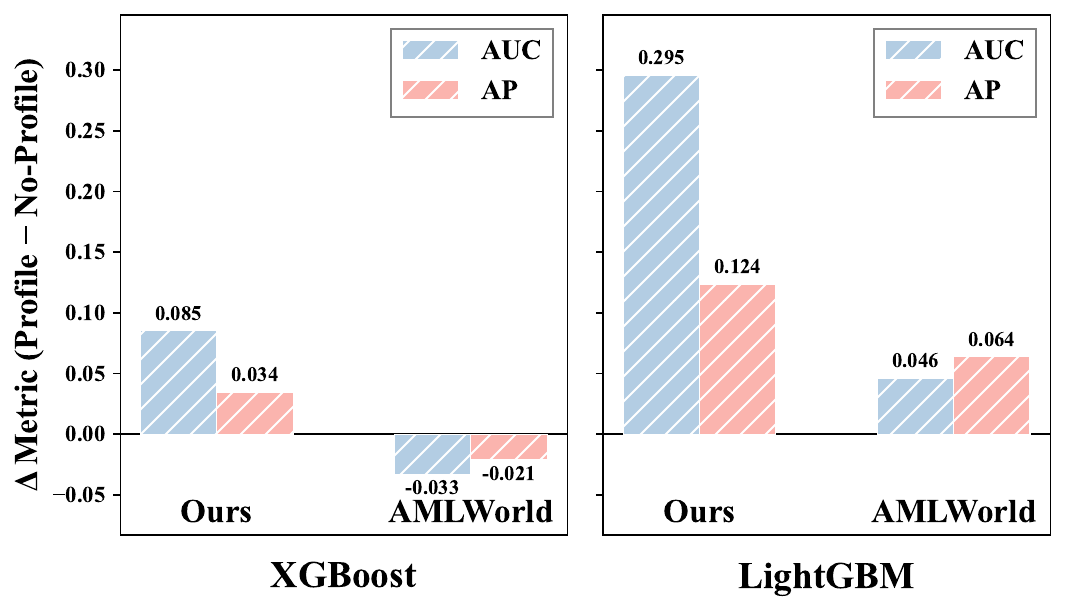}
    \caption{Change in AUC and AP after adding node profile features to GBT detectors on \textbf{TransXion} and \textbf{AMLWorld}.}
    \label{fig:barchart}
\end{figure}
\textit{Cross-benchmark gain comparison.}
We compare the marginal benefit of adding node-level context across benchmarks. Fig~\ref{fig:barchart} reports the change in ranking and retrieval performance after enabling profile features, measured by $\Delta$AUC and $\Delta$AP relative to profile-blind baselines on TransXion and AMLWorld. The gains are consistently larger on TransXion, with $\Delta$AUC $=0.085$ and $\Delta$AP $=0.034$ for XGBoost, and $\Delta$AUC $=0.295$ and $\Delta$AP $=0.124$ for LightGBM. In contrast, AMLWorld exhibits smaller and model-dependent effects, including slight degradations for XGBoost. This gap indicates that the entity profiles in TransXion provide richer and more predictive semantic context, enabling detectors to better disambiguate transactions that appear similar at the event level and supporting a more rigorous evaluation of semantic consistency reasoning.

\begin{table}[t]
\centering
\caption{Effect of profile-aware node features on GNN detectors on TransXion.}
\label{tab:profile_gnn}
\resizebox{\columnwidth}{!}{
\begin{tabular}{llrrrr}
\toprule
Setting & Metric & Multi-GIN & Multi-GIN+EU & Multi-PNA & Multi-GAT \\
\midrule
Edge-only & AUC & \textbf{0.9688} & 0.9528 & 0.9683 & \textbf{0.9516} \\
 & AP & 0.2878 & 0.2690 & 0.3508 & 0.2779 \\
 & F1 & 0.4340 & 0.4247 & \textbf{0.4865} & 0.4336 \\
\midrule
Profile+Edge & AUC & 0.9550 &\textbf{ 0.9718} & \textbf{0.9759} & 0.9452 \\
 & AP & \textbf{0.3173} & \textbf{0.3176} & \textbf{0.3746} & \textbf{0.2916} \\
 & F1 & \textbf{0.4666} & \textbf{0.4660} & 0.4732 & \textbf{0.4654} \\
\bottomrule
\end{tabular}
}
\end{table}
\textit{Profile-aware GNN detection.}
We further evaluate whether semantic profiles also benefit graph-based detectors. In the profile-aware setting, node attributes are fused with edge-level transaction features before edge classification. As shown in Table~\ref{tab:profile_gnn}, adding profile information improves AP for all evaluated GNN backbones and improves F1 for three out of four backbones. This confirms that TransXion's entity profiles provide useful semantic signals not only for tabular detectors, but also for relational models that jointly encode transaction topology and node context.

\begin{table}[t]
\centering
\caption{Ablation of synthesis strategies on the \textit{AMLWorld} dataset. For each detector, the \textbf{lowest} values are bolded and the \underline{second lowest} values are underlined}
\label{tab:amlworld_rl_ablation}
\fontsize{9.5pt}{10.5pt}\selectfont
\begin{tabular*}{\columnwidth}{@{\extracolsep{\fill}}ll cc@{}}
\toprule
\textbf{Configuration} & \textbf{Metric} & \textbf{Multi-GIN} & \textbf{XGBoost} \\
\midrule
\multirow{3}{*}{Original}      & AUC & 0.9864 & 0.9631 \\
                               & AP  & 0.6380 & \underline{0.3761} \\
                               & F1  & 0.6366 & \underline{0.4046} \\
\midrule
\multirow{3}{*}{LLM-Only}      & AUC & \underline{0.9841} & \underline{0.9481} \\
                               & AP  & \underline{0.5384} & 0.4008 \\
                               & F1  & \underline{0.5741} & 0.4573 \\
\midrule
\multirow{3}{*}{RL-Optimized}  & AUC & \textbf{0.9796} & \textbf{0.9474} \\
                               & AP  & \textbf{0.5121} & \textbf{0.3276} \\
                               & F1  & \textbf{0.5037} & \textbf{0.3849} \\
\bottomrule
\end{tabular*}
\end{table}

\subsubsection{Effect of Anomaly Synthesis Mechanism.}
To verify that the increased detection difficulty of our benchmark is not caused by stochastic noise or structural degradation, we conduct a controlled hardening experiment using AMLWorld as a fixed topological substrate. By keeping the underlying graph backbone constant, this design isolates the impact of our synthesis agent and decouples the effect of adversarial optimization from confounding factors such as dataset scale and inherent metadata distributions.

Notably, we deliberately select \textit{AMLWorld} for this validation rather than our own \textit{TransXion} backbone to ensure an impartial and objective evaluation. By demonstrating that our synthesis agent can successfully "harden" a well-established third-party benchmark, we prove that the observed hardness of TransXion is a universal property of our adversarial framework rather than an artifact of a specific synthetic data distribution. This controlled design, with the AMLWorld backbone held fixed, allows us to directly observe how the optimization procedure reshapes detection boundaries.

We evaluate three graph-level configurations that differ only in whether and how the benchmark has been modified. 
Original is the released AMLWorld graph in its unmodified form. 
LLM-Only is a hardened variant produced by an LLM-based synthesis agent without monitor-guided optimization.
RL-Optimized is a hardened variant produced by the same synthesis pipeline using an agent whose decision policy is obtained through monitor-guided training, after which the learned policy is used to generate edits.

As summarized in Table~\ref{tab:amlworld_rl_ablation}, the three configurations exhibit a clear progression in detection difficulty. The LLM-Only hardening produces mixed effects: it reduces Multi-GIN performance across metrics, yet it can improve some operating-point metrics for XGBoost, indicating that unguided hardening does not consistently increase difficulty across detector families. In contrast, the RL-Optimized variant yields a consistent degradation for both Multi-GIN and XGBoost, with lower scores across all reported metrics relative to the Original setting. This contrast suggests that monitor-guided optimization is necessary to produce systematic hardening, whereas LLM-only edits can shift the data distribution in ways that are not uniformly adversarial to different model classes. Overall, these results emphasize that the high detection difficulty of TransXion is a deliberate consequence of adaptive adversarial optimization rather than stochastic noise or structural degradation.

\textit{Cross-family transferability.}
To examine whether monitor-guided hardening overfits to a specific detector family, we further conduct a cross-family transfer experiment: the synthesis process is guided by a GBT-based monitor, while evaluation is performed using Multi-GIN. As shown in Table~\ref{tab:cross_monitor_transfer}, GBT-guided hardening still substantially degrades Multi-GIN performance, reducing AP from 0.6380 to 0.4297 and F1 from 0.6366 to 0.4807. This degradation is stronger than the LLM-Only variant, suggesting that the generated anomalies are not merely tailored to a specific GNN monitor, but encode transferable challenging patterns across detector families.

\begin{table}[t]
\centering
\caption{Cross-family transfer of hardened anomalies. Synthesis is guided by a GBT monitor and evaluated by Multi-GIN.}
\label{tab:cross_monitor_transfer}
\fontsize{9pt}{11pt}\selectfont
\setlength{\tabcolsep}{3pt}
\renewcommand{\arraystretch}{0.95}
\begin{tabular*}{\columnwidth}{@{\extracolsep{\fill}}lccc@{}}
\toprule
\textbf{Configuration} & \textbf{AUC} & \textbf{AP} & \textbf{F1} \\
\midrule
Original & 0.9864 & 0.6380 & 0.6366 \\
LLM-Only & 0.9841 & 0.5384 & 0.5741 \\
GBT-Guided & 0.9738 & 0.4297 & 0.4807 \\
\bottomrule
\end{tabular*}
\end{table}

\paragraph{Summary for RQ2.}
Across tabular and graph-based detectors, TransXion provides a more challenging and informative AML benchmark than prior public datasets. Under a unified edge-feature-only protocol, TransXion yields the weakest practical detection performance for both GBT and GNN detectors, especially in AP and F1, which are more relevant to alerting under extreme class imbalance. Adding entity profiles consistently improves detection performance, indicating that TransXion provides predictive semantic context beyond edge-level transaction statistics. Finally, the controlled hardening experiments on AMLWorld show that the increased difficulty is not merely caused by random noise or dataset scale, but is systematically induced by the monitor-guided non-template anomaly synthesis process. Together, these results show that TransXion shifts AML evaluation from shortcut recognition of fixed motifs toward context-aware reasoning over profiles, temporal behavior, and transaction topology.

%% file: 6_Conclusion.tex
We introduce TransXion, a high-fidelity transaction-graph benchmark designed to address two fundamental limitations of existing AML datasets: the lack of rich entity-level context and the reliance on template-driven anomaly injection. By jointly modeling persistent entity profiles, profile-conditioned normal behavior, and adaptive non-template anomaly synthesis, TransXion captures realistic interactions between transactional topology and semantic context that are central to operational AML.

Extensive fidelity analyses show that TransXion preserves key stylized properties of real payment networks, including connected backbones, hierarchical core--periphery structure, disassortative mixing, and economically plausible heavy-tailed behavior. At the same time, detection experiments demonstrate that TransXion is substantially more challenging than widely used benchmarks across both graph-based and feature-based models. The observed performance degradation reflects reduced shortcut learning and increased reliance on contextual reasoning, rather than artifacts of noise or synthetic complexity.

Overall, TransXion provides a more faithful and demanding testbed for developing, evaluating, and stress-testing AML detection methods. We hope this benchmark will support future research on context-aware, robust, and auditable transaction monitoring systems that better reflect the realities of financial crime detection.

%% file: 7_Appendix.tex
\section{Details of Anomalous Transaction Synthesis}
\label{Appendix:Anomalous}

\subsection{Scope and Disclosure Policy}
This appendix describes the reinforcement-learning (RL) training pipeline used for anomaly synthesis, including monitor-based feedback, group-relative optimization, and key training settings for reproducibility. To reduce misuse risk, we avoid disclosing institution-specific operational rules or case-identifying details, while reporting the model-agnostic optimization setup used to construct the benchmark.
For GRPO-based synthesis, we use AdamW with learning rate $1\times10^{-5}$ and weight decay $0.01$, gradient accumulation steps of $4$, and maximum gradient norm $1.0$. The policy is adapted with LoRA using rank $r=8$, $\alpha=16$, and dropout $0.05$. In our implementation, LoRA updates approximately $0.07\%$ of the base model parameters. This synthesis cost is incurred only once during benchmark construction; downstream users can directly use the released fixed benchmark for detection evaluation.

\subsection{System Overview}
The system consists of two components: (i) a policy that generates candidate graph-edit actions, and (ii) an external detector monitor that evaluates transaction subgraphs and outputs detection-related performance metrics. These metrics are aggregated into a scalar score, which serves as the learning signal for policy optimization. Training follows the Group Relative Policy Optimization (GRPO) paradigm, which estimates advantages using within-group statistics rather than an explicit value function.

\subsection{Environment Interaction and Group Sampling}
Let $C$ denote a transaction cluster represented as a subgraph. For each cluster, the policy $\pi_\theta$ samples $K$ trajectories,
\[
\{\tau_1,\ldots,\tau_K\} \sim \pi_\theta(\cdot \mid C).
\]
Each trajectory $\tau_i$ consists of a sequence of interaction steps. At step $t$, the agent observes a state derived from the current graph, emits a structured action proposing a graph edit, and the environment applies this edit to obtain the next graph state.

\subsection{Monitor Feedback and Reward Construction}
At each interaction step $t$, the monitor is applied to both the pre-action graph $G_t$ and the post-action graph $G_{t+1}$, yielding scalar scores:
\[
S_{\mathrm{pre}} = S(G_t), \qquad S_{\mathrm{post}} = S(G_{t+1}).
\]
The score is computed as a weighted combination of multiple detection metrics produced by the monitor, such as F1 score, area under the ROC curve (AUC), and average precision (AP). Formally,
\[
S = w_1 \cdot \mathrm{F1} + w_2 \cdot \mathrm{AUC} + w_3 \cdot \mathrm{AP},
\]
where $w_1,w_2,w_3$ are non-negative weights determined by the experimental configuration.

The step reward consists of two terms:
\[
R_t = R_{\mathrm{validity}}(a_t) + \lambda_{\mathrm{mon}} \cdot \Delta S_t,
\]
with
\[
\Delta S_t = S_{\mathrm{pre}} - S_{\mathrm{post}}.
\]
The validity term $R_{\mathrm{validity}}(a_t)$ penalizes malformed or non-executable structured actions and may terminate the trajectory in case of severe violations. The monitor-difference term $\Delta S_t$ provides feedback proportional to the reduction in the monitor's detection performance induced by the applied edit, scaled by the coefficient $\lambda_{\mathrm{mon}}$.

\subsection{Group-Relative Advantage Estimation}
GRPO estimates advantages using group-level statistics. Let $J_i$ denote the cumulative return of trajectory $\tau_i$. The group mean and standard deviation are computed as
\[
\mu = \frac{1}{K}\sum_{i=1}^K J_i, \qquad
\sigma = \sqrt{\frac{1}{K}\sum_{i=1}^K (J_i-\mu)^2}.
\]
The standardized advantage for trajectory $i$ is then defined as
\[
A_i = \frac{J_i - \mu}{\sigma + \epsilon},
\]
where $\epsilon>0$ is a small constant for numerical stability.

\subsection{Policy Optimization Objective}
The policy is optimized by maximizing the likelihood of actions or tokens associated with higher relative advantage. The resulting objective can be written as
\[
\mathcal{L}(\theta)
=
-\frac{1}{K}\sum_{i=1}^K A_i \sum_{t}\log \pi_\theta(x_{i,t}\mid x_{i,<t}, C).
\]
This formulation enables stable optimization without training a separate critic network, while retaining a relative-performance baseline within each group.
\section{Detailed Dataset Information}
\label{Appednix:detailed dataset}

\subsection{Dataset Overview}
\label{subsec:dataset_overview}

We use a synthetically generated transaction dataset for AML-related experiments. Each row represents a directed transaction from a sender to a receiver at a specific timestamp. The sender endpoint is identified by From Bank and From Account, and the receiver endpoint is identified by To Bank and To Account. Transaction values are recorded on both sides: Amount Paid in Payment Currency for the payer side, and Amount Received in Receiving Currency for the payee side, allowing paid and received amounts to differ under currency conversion. The transaction channel is captured by Payment Format. A binary label, is\_laundering, indicates whether a transaction is laundering (1) or benign (0).

For subsequent statistics, we construct a unique account key by combining bank and account identifiers to avoid cross-bank identifier collisions. When computing account activity, each transaction contributes one count to both the sender and receiver accounts. Since the dataset provides transaction-level laundering labels but not pattern-level typology identifiers, we report aggregate laundering prevalence and distributional characteristics without pattern-wise breakdowns.

\sisetup{
  group-separator = {,},
  group-minimum-digits = 4
}

\subsection{Dataset Scale and Label Statistics}
\label{subsec:dataset_scale}

Table~\ref{tab:dataset_stats} reports the dataset scale and the laundering label prevalence.
The dataset spans 365 days and contains 47{,}526 unique bank accounts and 3{,}029{,}170 transactions.
Among them, 4{,}641 transactions are labeled as laundering.
This corresponds to a prevalence of 0.153\% (approximately 1 in 653 transactions).
The resulting class imbalance is substantial, so downstream evaluations should emphasize ranking- and precision-oriented metrics (e.g., PR-AUC, precision@K) rather than accuracy.

\begin{table}[!t]
\centering
\small
\setlength{\tabcolsep}{7pt}
\renewcommand{\arraystretch}{1.15}
\caption{Dataset summary statistics}
\label{tab:dataset_stats}
\begin{tabular}{
  S[table-format=3.0]
  S[table-format=5.0]
  S[table-format=7.0]
  S[table-format=4.0]
  S[table-format=3.2]
}
\toprule
{Days} & {Accounts} & {Transactions} & {Laundering} & {$1$ per $N$} \\
\midrule
365 & 47526 & 3029170 & 4641 & 652.70 \\
\bottomrule
\end{tabular}
\end{table}

\begin{figure}[!t]
    \centering
    \includegraphics[width=1\linewidth]{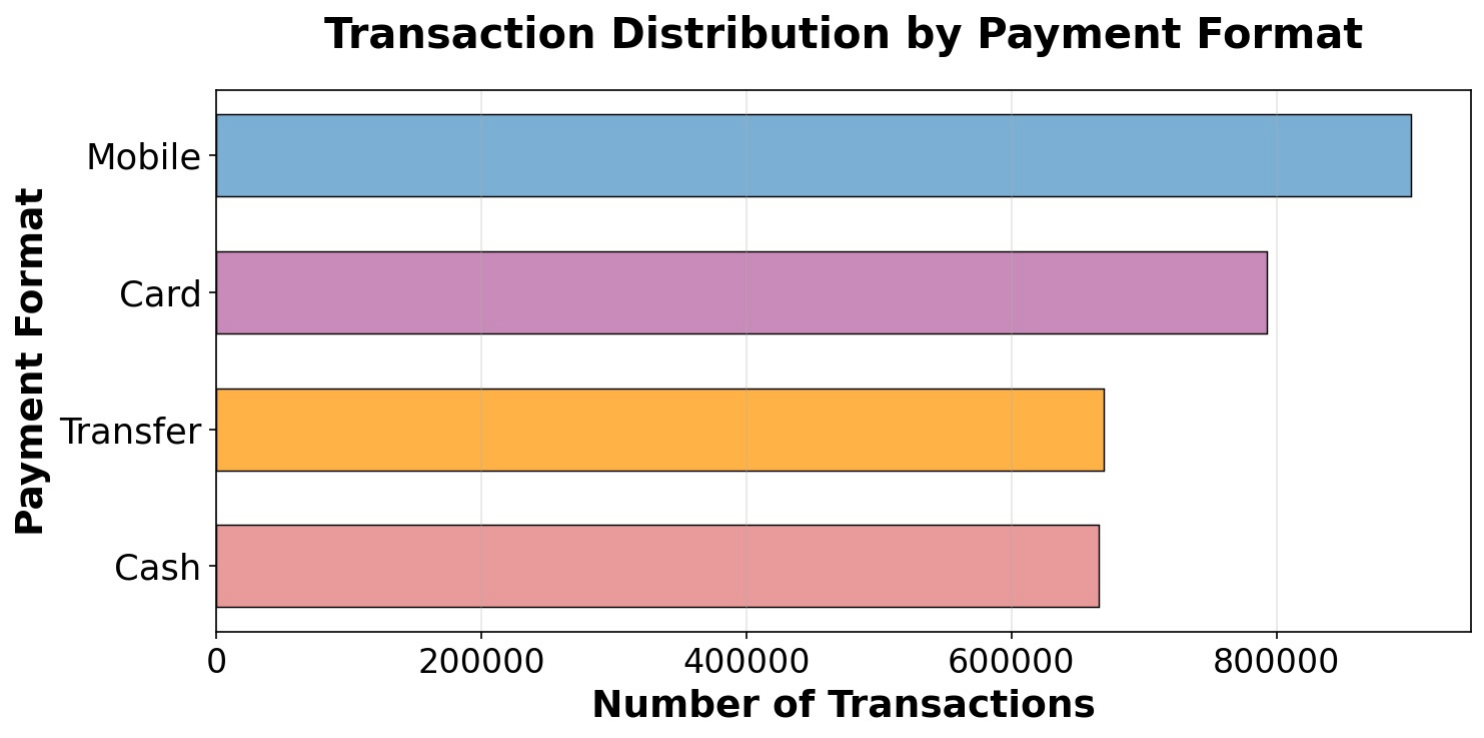}
    \caption{Transaction counts by payment format. Bars report the number of transactions in each \texttt{Payment Format} category.}
    \label{fig:payment_format}
\end{figure}

\subsection{Transaction-Level Characteristics}
\label{subsec:transaction_level}

Figure~\ref{fig:payment_format} summarizes the transaction composition by \texttt{Payment Format}.
Mobile payments account for the largest share, followed by card payments, while transfers and cash are less frequent.
This skewed mix implies that channel information can be a strong cue in modeling.
To avoid learning trivial shortcuts, we later report results with metrics that are robust to class imbalance and recommend stratified analyses by payment format when comparing methods.

\begin{table*}[!t]
\centering
\caption{Tail-fit diagnostics under automatic $x_{\min}$ selection. $D$ is the KS distance; $(x_{\min}\mid n_{\text{tail}})$ reports the selected threshold and tail size; $(R,p)$ compares power law against lognormal.}
\label{tab:tail-diagnostics-appendix}
\fontsize{7pt}{9pt}\selectfont
\setlength{\tabcolsep}{3pt}
\renewcommand{\arraystretch}{0.9}
\begin{tabular*}{\textwidth}{@{\extracolsep{\fill}}llcccc@{}}
\toprule
\textbf{Dataset} & \textbf{Distribution} &
\textbf{$x_{\min}\mid n_{\text{tail}}$} &
\textbf{$\alpha$} & \textbf{$D$} &
\textbf{$R,p$} \\
\midrule
Ours & degree\_unique    & 56 $\mid$ 24.0k     & 2.904 & 0.190 & $-4.34{\times}10^{2}$, $4.92{\times}10^{-30}$ \\
Ours & degree\_tx\_count & 79 $\mid$ 23.8k     & 2.904 & 0.108 & $-3.88$, $6.25{\times}10^{-1}$ \\
Ours & strength          & 120.8k $\mid$ 20.0k & 2.405 & 0.057 & $-1.12$, $1.48{\times}10^{-1}$ \\
Ours & amount            & 40.3k $\mid$ 30.3k  & 2.926 & 0.042 & $-3.26{\times}10^{1}$, $3.11{\times}10^{-3}$ \\
\midrule
AMLWorld & degree\_unique    & 4 $\mid$ 111.3k       & 3.000 & 0.045 & $-1.79{\times}10^{3}$, $1.04{\times}10^{-80}$ \\
AMLWorld & degree\_tx\_count & 40 $\mid$ 79.6k       & 3.000 & 0.049 & $-2.97{\times}10^{3}$, $1.82{\times}10^{-87}$ \\
AMLWorld & strength          & 531.0M $\mid$ 5.2k    & 1.788 & 0.020 & $-4.23$, $6.84{\times}10^{-2}$ \\
AMLWorld & amount            & 623.8k $\mid$ 253.9k  & 1.542 & 0.022 & $-1.32{\times}10^{3}$, $2.01{\times}10^{-202}$ \\
\midrule
AMLSim & degree\_unique    & 12 $\mid$ 4.5k      & 2.931 & 0.044 & $-5.96{\times}10^{1}$, $6.42{\times}10^{-11}$ \\
AMLSim & degree\_tx\_count & 219 $\mid$ 4.9k     & 2.968 & 0.030 & $-2.23{\times}10^{1}$, $3.86{\times}10^{-5}$ \\
AMLSim & strength          & 21.5M $\mid$ 4.4k   & 2.148 & 0.105 & $-2.56{\times}10^{2}$, $2.47{\times}10^{-39}$ \\
AMLSim & amount            & 85.5k $\mid$ 26.5k  & 1.327 & 0.158 & $-5.56{\times}10^{3}$, $<10^{-300}$ \\
\midrule
SAML-D & degree\_unique    & 1 $\mid$ 855.5k       & 3.000 & 0.118 & $-3.52{\times}10^{4}$, $<10^{-300}$ \\
SAML-D & degree\_tx\_count & 215 $\mid$ 17.1k      & 2.235 & 0.240 & $-5.77{\times}10^{3}$, $<10^{-300}$ \\
SAML-D & strength          & 80.6k $\mid$ 402.1k   & 2.228 & 0.045 & $1.57{\times}10^{1}$, $7.05{\times}10^{-308}$ \\
SAML-D & amount            & 10.2k $\mid$ 2471.3k  & 3.000 & 0.030 & $-2.19{\times}10^{2}$, $4.54{\times}10^{-23}$ \\
\bottomrule
\end{tabular*}
\end{table*}

\subsection{Account-Level Characteristics}
\label{subsec:account_level}

\begin{figure}
    \centering
    \includegraphics[width=1\linewidth]{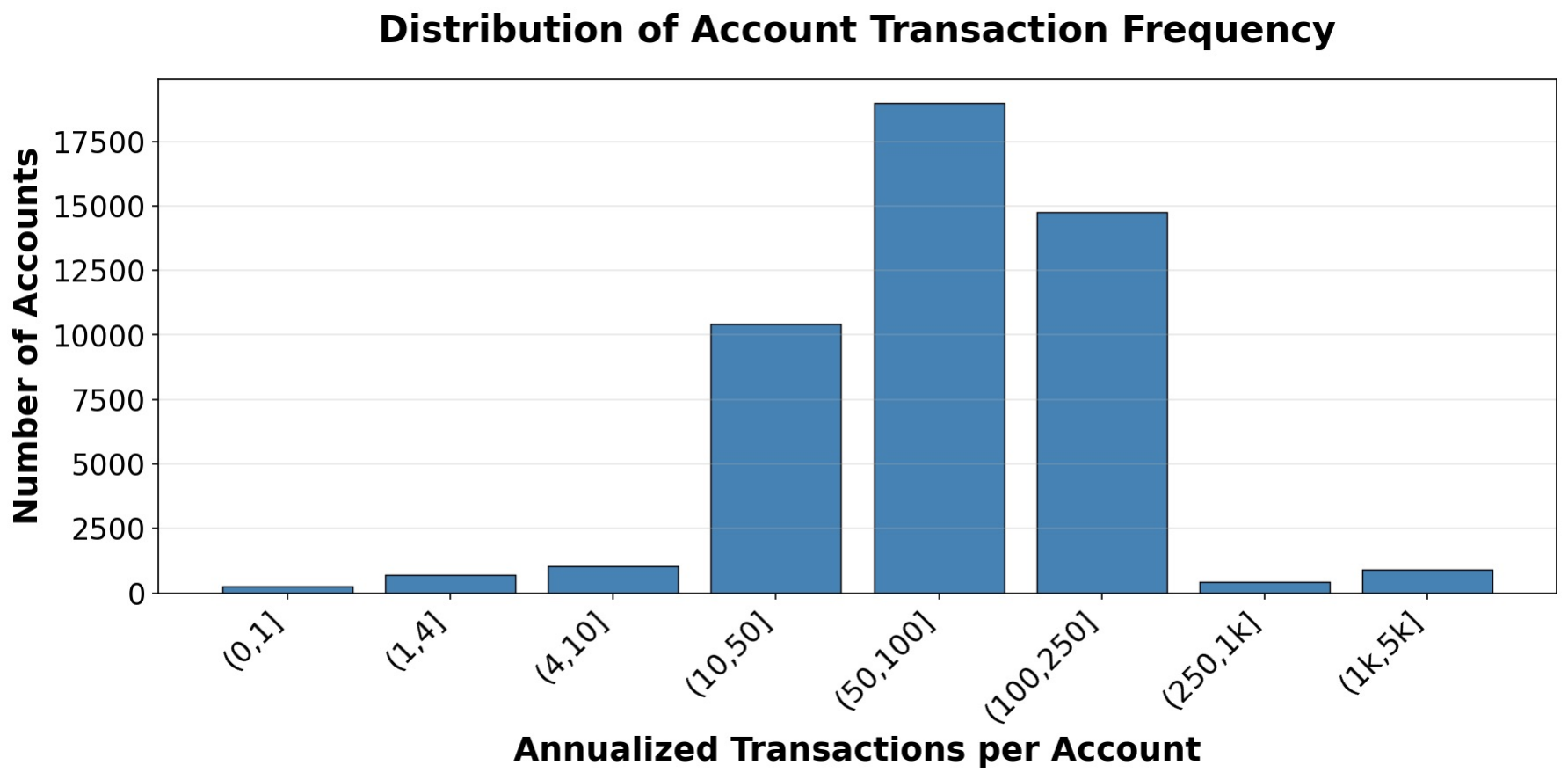}
    \caption{Annualized transaction frequency per account. Each account's incoming and outgoing transactions are counted and scaled to a yearly rate; bars report the number of accounts in each bin.}
    \label{fig:account_freq}
\end{figure}

Figure~\ref{fig:account_freq} reports the distribution of annualized transaction frequency per account.
We define an account as a bank--account pair and count both incoming and outgoing transactions.
Most accounts fall in the mid-frequency bins (10--250 transactions per year), while a small fraction is either rarely active or highly active.
This heterogeneity implies a skewed degree distribution in the induced transaction graph, which can affect both sampling and evaluation.
Accordingly, we later interpret performance with care for activity-driven bias, rather than treating all accounts as equally informative.

\section{Detailed Evidence of Fidelity}
\label{Appendix:Fidelity}

Table~\ref{tab:tail-diagnostics-appendix} reports tail-fit diagnostics under automatic $x_{\min}$ selection. These diagnostics are intended to complement the main graph-invariant analysis rather than to claim that a single metric universally determines realism. In particular, we examine whether tail regions are non-degenerate, whether fitted exponents vary across different observables, and whether structural and monetary variables exhibit distinct regimes. Under these criteria, the results are most consistent with \emph{Ours} exhibiting coherent and non-degenerate tail behavior across both structural and monetary variables.

\textit{Coherent and Non-Templated Tails.}
A key realism signal is that different observables should not collapse to a single universal exponent. In \emph{Ours}, the fitted $\alpha$ varies across structural and monetary variables: structural quantities such as degree\_tx\_count and strength remain in plausible heavy-tail regimes, while monetary amount follows a different tail pattern. In contrast, several baseline settings repeatedly return boundary values such as $\alpha=3.0000$, suggesting more rigid or over-regularized tail behavior rather than organically emerging heterogeneity.

\textit{Automatic Tail Selection.}
Under automatic $x_{\min}$ selection, \emph{Ours} achieves small KS distances for degree\_tx\_count ($D=0.1083$), strength ($D=0.0570$), and amount ($D=0.0418$). These values indicate that, once the tail region is selected in a data-adaptive manner, the empirical tails of \emph{Ours} are well captured by the fitted model. We do not interpret the smallest KS value alone as the sole criterion of realism; instead, we consider it together with exponent stability, non-degenerate tail sizes, and the distinction between structural and monetary variables.

\textit{Non-Degenerate Tail Definitions.}
Some baselines produce degenerate or weakly informative tail regions. For example, SAML-D degree\_unique selects $x_{\min}=1$ with $n_{\text{tail}}=n$, making the tail effectively equal to the full distribution. This weakens the interpretability of the fitted tail because no meaningful separation exists between the body and the tail. In contrast, \emph{Ours} retains meaningful tail sizes across variables, with selected tails that cover neither the entire support nor an extremely small residual subset.

\textit{Mixed Monetary and Structural Regimes.}
For monetary amount, \emph{Ours} favors a lognormal-like regime over a strict power law, while structural variables preserve credible heavy-tail signatures. This mixed behavior is plausible for payment systems: account activity and transaction volume can exhibit heavy-tailed participation, whereas transaction amounts often arise from multiplicative or mixture mechanisms rather than a pure power law. Therefore, the fidelity evidence is not that every observable follows the same distributional law, but that TransXion preserves heterogeneous tail behavior across semantically different variables.